\documentclass[11pt]{article}

\usepackage[preprint]{acl}
\usepackage{algorithm}
\usepackage{algpseudocode}
\usepackage{times}
\usepackage{amsmath}
\usepackage{latexsym}
\usepackage{amsmath}   
\usepackage{amssymb}   
\usepackage{xcolor}
\usepackage{color}
\usepackage{booktabs}
\usepackage{caption}
\usepackage{cuted}      
\usepackage{capt-of}    
\usepackage{booktabs}   
\usepackage{lipsum}     

\usepackage[T1]{fontenc}

\usepackage[utf8]{inputenc}

\usepackage{microtype}

\usepackage{inconsolata}

\usepackage{graphicx}

\title{Rewarding Creativity: A Human-Aligned Generative Reward Model for Reinforcement Learning in Storytelling}

\author{
    \textbf{Zhaoyan Li} \quad
    \textbf{Hang Lei}\thanks{Corresponding author.}\quad
    \textbf{Yujia Wang} \quad
    \textbf{Lanbo Liu} \quad
    \textbf{Hao Liu}\quad
    \textbf{Liang Yu}
    \\[1ex] 
    Alibaba Group
    \\[1ex]
    \texttt{\{lzy434483,leihang.lh,jessie.wyj,llb140195,lh414475,deyi.yl\}@alibaba-inc.com}
}

\begin{document}
\maketitle
\begin{abstract}
While Large Language Models (LLMs) can generate fluent text, producing high-quality creative stories remains challenging. Reinforcement Learning (RL) offers a promising solution but faces two critical obstacles: designing reliable reward signals for subjective storytelling quality and mitigating training instability.
This paper introduces the Reinforcement Learning for Creative Storytelling (RLCS) framework to systematically address both challenges. First, we develop a Generative Reward Model (GenRM) that provides multi-dimensional analysis and explicit reasoning about story preferences, trained through supervised fine-tuning on demonstrations with reasoning chains distilled from strong teacher models, followed by GRPO-based refinement on expanded preference data. Second, we introduce an entropy-based reward shaping strategy that dynamically prioritizes learning on confident errors and uncertain correct predictions, preventing overfitting on already-mastered patterns.
Experiments demonstrate that GenRM achieves 68\% alignment with human creativity judgments, and RLCS significantly outperforms strong baselines including Gemini-2.5-Pro in overall story quality. This work provides a practical pipeline for applying RL to creative domains, effectively navigating the dual challenges of reward modeling and training stability.
\end{abstract}

\section{Introduction}
Reinforcement Learning (RL) has driven significant advancements for Large Language Models (LLMs) in objective domains like mathematics and code generation, where verifiable reward functions provide clear optimization targets \citep{jain2025multiturncodegenerationsinglestep}. However, this paradigm does not readily translate to subjective, open-ended domains like creative writing, where the concept of a singular "ground truth" is meaningless. This ambiguity makes designing reliable reward signals a formidable challenge, creating a fundamental bottleneck for high-quality story generation.

Applying RL to storytelling faces two main obstacles. First, the \textbf{reward modeling problem}: existing solutions like LLM-based judges suffer from superficial biases \citep{wang2023voyageropenendedembodiedagent,feuer2025judgmentnoisedesignfailures}, while conventional discriminative reward models produce only scalar scores that fail to capture storytelling's multi-dimensional nature or provide explicit reasoning. Second, \textbf{training instability}: standard RL algorithms often suffer from policy collapse or reward hacking in text generation's vast action space, exploiting superficial patterns rather than developing genuine creative capabilities.

We introduce the \textbf{Reinforcement Learning for Creative Storytelling (RLCS)} framework to systematically address both challenges. First, our \textbf{Generative Reward Model (GenRM)} provides multi-dimensional analysis and explicit reasoning about story quality, articulating \textit{why} one story is preferred by evaluating aspects like plot coherence, character development, and creative originality. GenRM is trained through a two-stage pipeline: \textbf{supervised fine-tuning (SFT)} on demonstrations with reasoning chains from strong teacher models, followed by \textbf{GRPO-based refinement} on expanded preference data combining human annotations and multi-model consensus. This achieves 68\% agreement with human judges, substantially improving over traditional approaches.

Second, we introduce an \textbf{entropy-based reward shaping strategy} that dynamically prioritizes learning based on model confidence and correctness. Our approach focuses on (1) \textit{confident errors}, revealing systematic biases requiring correction, and (2) \textit{uncertain correct predictions}, indicating emerging capabilities needing reinforcement, while reducing emphasis on already-mastered patterns. This significantly improves training stability and convergence compared to uniform updates.

Our RLCS framework demonstrates substantial improvements over strong baselines. Stories generated by our model significantly outperform Gemini-2.5-Pro and other competitive systems, validating both the effectiveness of articulated rewards for capturing creative preferences and the importance of strategic optimization for stable RL training in subjective domains.

Overall, our contributions are threefold:
\begin{itemize}
    \item \textbf{Generative Reward Model with Explicit Reasoning:} We propose GenRM, which provides multi-dimensional analysis and explicit reasoning about story preferences rather than scalar scores. Trained through supervised fine-tuning and reinforcement learning, GenRM achieves 68\% alignment with human judgments, establishing a reliable foundation for RL in subjective domains.
    
    \item \textbf{Entropy-Based Reward Shaping:} We introduce a dynamic reward shaping strategy that prioritizes confident errors and uncertain correct predictions while preventing overfitting. This approach ensures stable and efficient policy optimization in creative generation.
    
    \item \textbf{Comprehensive RL Framework:} RLCS integrates articulated reward modeling with targeted optimization, providing a practical methodology for RL in subjective creative tasks. Our framework significantly outperforms strong baselines including Gemini-2.5-Pro.
\end{itemize}

Beyond offering a practical solution for creative writing, this work establishes a generalizable methodology for applying RL to complex, subjective generation tasks, providing a framework for more effective reinforcement learning in open-ended language generation domains.

\section{Related Work}

\paragraph{Creative Writing Evaluation.}
Evaluating creative writing remains challenging due to inherent subjectivity \cite{app15062971}. Traditional metrics like BLEU \cite{10.3115/1073083.1073135}, ROUGE \cite{lin-2004-rouge}, and METEOR \cite{banerjee-lavie-2005-meteor} measure n-gram overlap, failing to capture narrative coherence, originality, and creativity \cite{liu2023gevalnlgevaluationusing}. While human evaluation remains the gold standard, employing either crowd-workers \cite{chhun2024languagemodelsenjoystories, xie2023chapterstudylargelanguage} or domain experts \cite{chakrabarty2024artartificelargelanguage}, it is expensive, time-consuming, and suffers from low inter-annotator agreement \cite{marco2025readermetrictextualfeatures, app15062971}.

Recent work has shifted to using LLMs as automated evaluators. General-purpose models like GPT-4 \cite{bai2024longwriterunleashing10000word, wegmann-etal-2022-author} can be unreliable and biased \cite{chakrabarty2024artartificelargelanguage, chakrabarty2025aiwritingsalvagedmitigating, app15062971}. Specialized reward models like StoryER \cite{chen-etal-2022-storyer}, LitBench \cite{fein2025litbenchbenchmarkdatasetreliable}, and WritingBench \cite{xu2025largereasoningmodelssurvey} provide scalar scores or binary preferences but fail to explain \textit{why} one story is better. Our GenRM addresses this limitation by providing articulated, natural language feedback with explicit reasoning, offering richer guidance for reinforcement learning.

\paragraph{RL with Non-Verifiable Rewards.}
While RL excels in verifiable domains with automated feedback like mathematics and coding \citep{deepseekai2025deepseekr1incentivizingreasoningcapability, albalak2025bigmathlargescalehighqualitymath, chen2025r1codeinterpreterllmsreasoncode}, subjective tasks like creative writing lack ground-truth answers, making reward design challenging \citep{zhang2025surveyreinforcementlearninglarge}. The LLM-as-Judge paradigm \citep{zheng2023judgingllmasajudgemtbenchchatbot} has emerged as a solution, evolving through reasoning reward models \citep{li2023generativejudgeevaluatingalignment, ankner2024critiqueoutloudrewardmodels, chen2025rmr1rewardmodelingreasoning}, rubric-based methods \citep{jia2025writingzerobridgegapnonverifiable, gunjal2025rubricsrewardsreinforcementlearning, huang2025reinforcementlearningrubricanchors}, and co-evolving systems with self-rewarding \citep{yuan2025selfrewardinglanguagemodels, zhang2025critiquegrpoadvancingllmreasoning} or co-optimization strategies \citep{wang2025adaptivethinkingmodepolicy, hong2025coopercooptimizingpolicyreward}.

\paragraph{Generative Reward Models.}  
Unlike discriminative models \citep{cai2024internlm2technicalreport, yuan2024advancingllmreasoninggeneralists}, generative reward models leverage LLMs' generative capabilities for evaluation. Approaches include using general \citep{zheng2023judgingllmasajudgemtbenchchatbot} or specialized models \citep{li2023generativejudgeevaluatingalignment, cao2024compassjudger1allinonejudgemodel} as judges, extracting next-token probabilities as scores \citep{mahan2024generativerewardmodels, zhang2025generativeverifiersrewardmodeling}, or iterative training with synthetic preferences \citep{yuan2025selfrewardinglanguagemodels, wu2024metarewardinglanguagemodelsselfimproving}. These models can integrate with CoT \citep{kojima2023largelanguagemodelszeroshot} and RAG \citep{lewis2021retrievalaugmentedgenerationknowledgeintensivenlp}, enabling broader applications.

\section{Preliminaries}
\textbf{Notation}  In this paper, an autoregressive language model parameterized by $\theta$ is defined as a policy $\pi_{\theta}$. We use $x$ to denote a query and $D$ as the query set. Given a response $y$ to a query $x$, its likelihood under the policy $\pi_{\theta}$ is denoted as $\pi_{\theta}(y|x)=\prod_{t=1}^{|y|}\pi_{\theta}(y_t|x,y_{<t})$, where $|y|$ denotes the number of tokens in $y$. A query-response pair $(x, y)$ can be scored by a verifier $r$, resulting in a reward $r(x, y) \in [-1, 1]$.

\textbf{Group Relative Policy Optimization (GRPO)} GRPO \citep{shao2024deepseekmathpushinglimitsmathematical} bypasses the need for the value model by computing the relative advantage of each response within a group of responses to the same query. Specifically, GRPO optimizes the following objective:
\begin{equation}
\label{eq:grpo}
\begin{aligned}
&\mathcal{J}_\text{GRPO}(\theta)=\mathbb{E}_{x\sim D,\{y_i\}_{i=1}^{G}\sim \pi_{\theta_{\text{old}}}}\bigg[\frac{1}{G}\sum_{i=1}^{G}\frac{1}{|y_i|}\sum_{t=1}^{|y_i|}\\
&\min\Big(\omega_{i,t}(\theta)A_{i,t}, \text{clip}(\omega_{i,t}(\theta), 1-\varepsilon, 1+\varepsilon)A_{i,t} \Big)\bigg]\\
\end{aligned}
\end{equation}
where $G$ is the number of generated responses to each query $x$, and the importance ratio $\omega_{i,t}(\theta)$ and advantage $A_{i,t}$ of token $y_{i,t}$ are defined as:
\begin{equation}
\omega_{i,t}=\frac{\pi_{\theta}(y_{i,t}|x,y_{i,<t})}{\pi_{\theta_{\text{old}}}(y_{i,t}|x,y_{i,<t})}
\end{equation}

\begin{equation}
A_{i,t}=\frac{r(x,y_i)-\text{mean}(\{r(x,y_i)\}_{i=1}^{G})}{\text{std}(\{r(x,y_i)\}_{i=1}^{G})}
\end{equation}

\section{Constructing an Articulated Reward Model}
Unlike traditional discriminative reward models that output a single scalar score, the Generative Reward Model (GenRM) provides structured, multi-dimensional feedback by explicitly reasoning about \emph{why} one story is preferred over another. This richer feedback signal proves crucial for guiding the downstream story generation policy.

The training of GenRM proceeds in two stages: (1) A supervised fine-tuning (SFT) cold-start phase that establishes basic task-following and generation capabilities, and (2) A RL phase that continuously improves judgment accuracy through alignment with human preferences and consensus from strong teacher models.

\begin{figure*}[t]
    \centering
    \includegraphics[width=\textwidth]{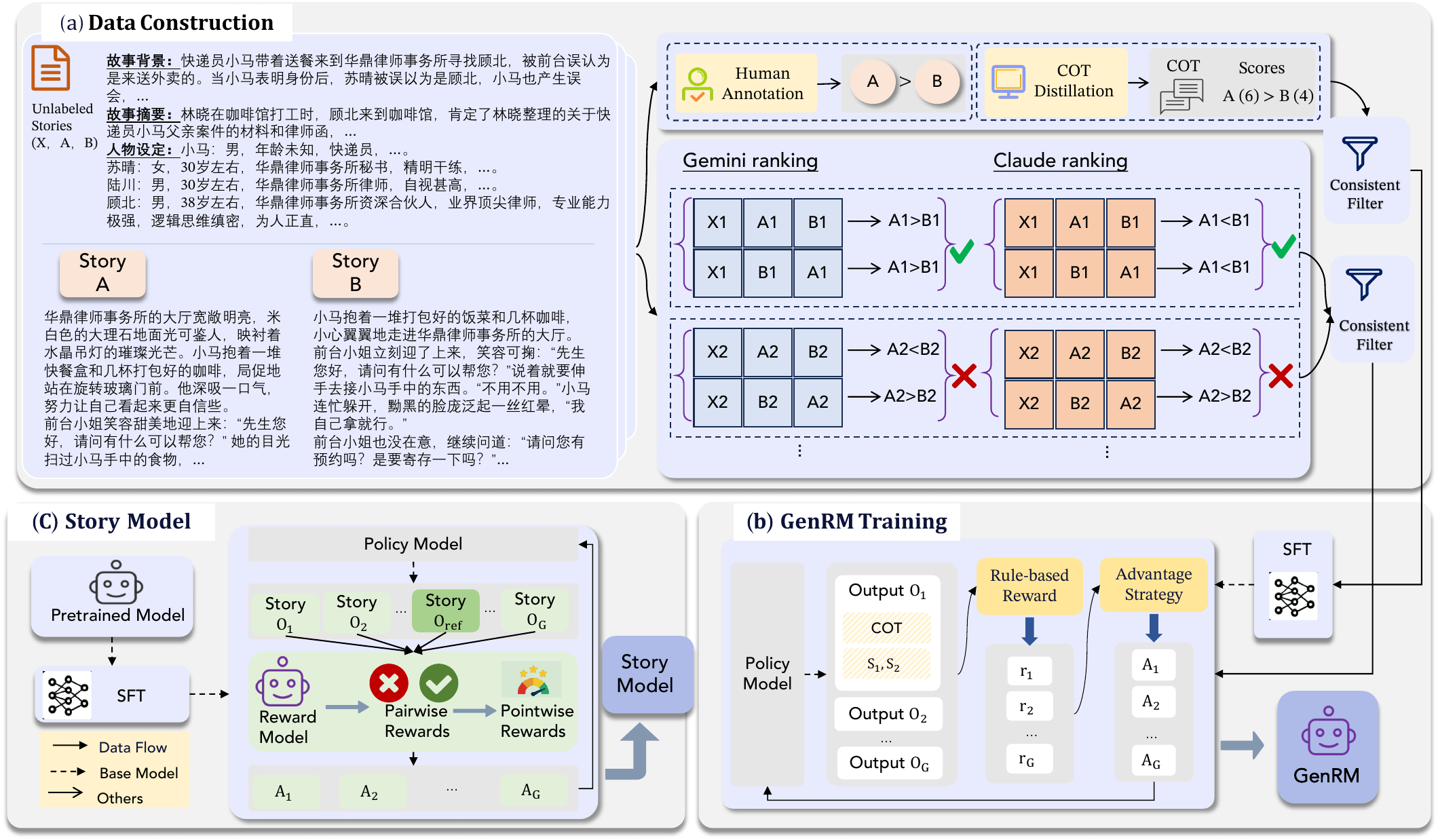} 
    \caption{Overview of the RLCS Framework: (a) Data Construction from unlabeled stories and human feedback, (b) Generative Reward Model (GenRM) training with rule-based rewards, and (c) Story Model training via supervised fine-tuning and reinforcement learning with GenRM guidance.}
    \label{fig:Overview_RLCS}
\end{figure*}

\subsection{Task Formulation}
We begin by formally defining the story generation task that our framework aims to optimize. Given story context $\mathbf{c} = \{p, h, o\}$, where $p$ denotes character profiles detailing personalities, relationships, and motivations, $h$ represents previous plot developments that precede the current generation point, and $o$ specifies the outline with key events and narrative goals for the next segment, the objective is to generate a story continuation $\mathbf{s}$ that: (1) maintains consistency with character profiles $p$, (2) closely follows the specified outline $o$, and (3) provides natural and coherent progression from the previous plot $h$.

To guide the story generation policy toward high-quality outputs, we introduce a Generative Reward Model (GenRM) that addresses the core challenge of providing reliable, interpretable reward signals for creative writing. As illustrated in Figure~\ref{fig:Overview_RLCS}(b), given the same story context $\mathbf{c}$ and two candidate continuations $\mathbf{s}_1$ and $\mathbf{s}_2$, GenRM is trained to perform two interconnected tasks. First, GenRM generates chain-of-thought analysis $\mathbf{d}$ that articulates a structured evaluation decomposing story quality into interpretable dimensions, including narrative coherence, creative originality, emotional engagement, and context-dependent criteria such as outline adherence and character consistency. Second, GenRM determines preference by synthesizing the multi-dimensional analysis to produce a justified preference judgment $y \in \{\mathbf{s}_1 \succ \mathbf{s}_2, \mathbf{s}_2 \succ \mathbf{s}_1\}$ that explicitly explains \textit{why} one continuation is superior based on the identified quality dimensions.

\subsection{Task Alignment through Supervised Fine-Tuning}
Before applying reinforcement learning, the model must first acquire foundational capabilities in task comprehension and structured output generation. As shown in Figure~\ref{fig:Overview_RLCS}(a), the cold-start stage addresses this need through supervised learning on high-quality demonstrations with complete reasoning chains.

\noindent\textbf{Human Annotation Data.} 
We begin with a dataset $\mathcal{D}_{\text{human}} = \{(\mathbf{c}_i, \mathbf{s}_1^i, \mathbf{s}_2^i, y_i^*)\}_{i=1}^{N}$ consisting of over 4,000 story pairs annotated by professional screenwriters, where each sample includes story context $\mathbf{c}_i$, two candidate stories $(\mathbf{s}_1^i, \mathbf{s}_2^i)$, and a preference label $y_i^* \in \{\mathbf{s}_1^i \succ \mathbf{s}_2^i, \mathbf{s}_2^i \succ \mathbf{s}_1^i\}$ indicating which story is better. Crucially, these annotations contain only the final judgment without the underlying reasoning process that explains \textit{why} one story is preferred over the other.

\noindent\textbf{Chain-of-Thought Distillation.} 
To construct training data with complete reasoning chains, we employ a distillation approach using Gemini-2.5-Pro as the teacher model $\mathcal{M}_{\text{teacher}}$, as depicted in the “COT Distillation” component of Figure~\ref{fig:Overview_RLCS}(a). For each story pair, the teacher model first analyzes the characteristics of both stories across multiple dimensions, then provides scores and a final preference judgment. The process consists of three key steps:

\begin{itemize}
    \item \textbf{CoT Generation}: For each sample, we prompt the teacher model to generate both a detailed reasoning chain $\mathbf{d}_i$ and a preference judgment $\hat{y}_i$:
    \begin{equation}
        (\mathbf{d}_i, \hat{y}_i) \sim \mathcal{M}_{\text{teacher}}(\mathbf{c}_i, \mathbf{s}_1^i, \mathbf{s}_2^i)
    \end{equation}
    
    \item \textbf{Position Bias Mitigation}: To address position bias in LLMs, we generate judgments for both the original and swapped story orders. Specifically, we obtain predictions in both presentation orders:
    \begin{align}
        (\mathbf{d}_i^{\text{orig}}, \hat{y}_i^{\text{orig}}) &\sim \mathcal{M}_{\text{teacher}}(\mathbf{c}_i, \mathbf{s}_1^i, \mathbf{s}_2^i) \\
        (\mathbf{d}_i^{\text{swap}}, \hat{y}_i^{\text{swap}}) &\sim \mathcal{M}_{\text{teacher}}(\mathbf{c}_i, \mathbf{s}_2^i, \mathbf{s}_1^i)
    \end{align}
    
    \item \textbf{Consistency Filtering}: We retain only samples where the teacher model's predictions are (1) consistent across both presentation orders and (2) aligned with human annotations:

    \begin{equation}
    \begin{aligned}
        \mathcal{D}_{\text{SFT}} = \{&(\mathbf{c}_i, \mathbf{s}_1^i, \mathbf{s}_2^i, \mathbf{d}_i, y_i^*) : \\
        &\hat{y}_i^{\text{orig}} = \hat{y}_i^{\text{swap}} = y_i^*\}
    \end{aligned}
    \end{equation}

\end{itemize}

Through this careful filtering process, we obtain $|\mathcal{D}_{\text{SFT}}| \approx 1{,}400$ high-quality training samples with verified reasoning chains. The supervised fine-tuning objective maximizes the likelihood of generating both the reasoning process and the correct preference:
\begin{equation}
    \mathcal{L}_{\text{SFT}} = -\mathbb{E}_{(\mathbf{c}, \mathbf{s}_1, \mathbf{s}_2, \mathbf{d}, y^*) \sim \mathcal{D}_{\text{SFT}}} [\log p_{\theta}(\mathbf{d}, y^* | \mathbf{c}, \mathbf{s}_1, \mathbf{s}_2)]
\end{equation}

This cold-start stage, represented by the SFT component in Figure~\ref{fig:Overview_RLCS}(b), equips GenRM with essential capabilities for subsequent reinforcement learning: understanding the evaluation task, generating structured multi-dimensional reasoning, and producing consistent judgments.

\subsection{Training the Generative Reward Model}
While the cold-start phase establishes basic competence in task understanding and structured reasoning generation, the GenRM's judgment accuracy must be further refined to align closely with human preferences. As illustrated in Figure~\ref{fig:Overview_RLCS}(b), reinforcement learning provides a natural framework for this optimization.

The samples from $\mathcal{D}_{\text{human}}$ that were filtered out during cold-start (due to inconsistent teacher model predictions or position bias) now serve as the initial RL training set:
\begin{equation}
    \mathcal{D}_{\text{RL}}^{\text{human}} = \mathcal{D}_{\text{human}} \setminus \mathcal{D}_{\text{SFT}}
\end{equation}
This design ensures efficient data utilization: high-quality samples with reliable reasoning chains are used for supervised learning, while the remaining samples with gold human labels guide policy improvement through reinforcement learning.

However, professional annotations are expensive and limited in scale. To address this challenge, we construct synthetic preference data through consensus among multiple state-of-the-art LLMs $\{\mathcal{M}_1, \mathcal{M}_2, \ldots, \mathcal{M}_K\}$ (e.g., Gemini-2.5-Pro, Claude-Sonnet-4), as shown in the “Consistent Filter” component of Figure~\ref{fig:Overview_RLCS}(a). For each story pair $(\mathbf{c}_i, \mathbf{s}_1^i, \mathbf{s}_2^i)$, we obtain judgments from all teacher models for both the original and swapped orderings:
\begin{equation}
    y_{i,k}^{\text{orig}} \sim \mathcal{M}_k(\mathbf{c}_i, \mathbf{s}_1^i, \mathbf{s}_2^i), \quad 
    y_{i,k}^{\text{swap}} \sim \mathcal{M}_k(\mathbf{c}_i, \mathbf{s}_2^i, \mathbf{s}_1^i)
\end{equation}
for $k = 1, \ldots, K$. We retain only samples where all models agree and demonstrate position-invariant judgments:

\begin{equation}
\mathbb{I}_{\text{consist}}^{(i)} = \mathbb{I}\left[\bigwedge_{k \neq k'} \left(y_{i,k}^{\text{orig}} = y_{i,k}^{\text{swap}} = y_{i,k'}^{\text{orig}} = y_{i,k'}^{\text{swap}}\right)\right]
\end{equation}

The filtered synthetic dataset is:
\begin{equation}
    \mathcal{D}_{\text{RL}}^{\text{syn}} = \left\{(\mathbf{c}_i, \mathbf{s}_1^i, \mathbf{s}_2^i, y_i) : \mathbb{I}_{\text{consist}}^{(i)} = 1\right\}
\end{equation}
where $y_i$ denotes the agreed-upon preference. The final RL training set combines both sources: $\mathcal{D}_{\text{RL}} = \mathcal{D}_{\text{RL}}^{\text{human}} \cup \mathcal{D}_{\text{RL}}^{\text{syn}}$.

We employ GRPO to refine the GenRM. For each training sample $(\mathbf{c}_i, \mathbf{s}_1^i, \mathbf{s}_2^i, y_i^*)$ from $\mathcal{D}_{\text{RL}}$, we denote the input as $\mathbf{x}_i = (\mathbf{c}_i, \mathbf{s}_1^i, \mathbf{s}_2^i)$ and sample $G$ independent outputs from the current policy:
\begin{equation}
    \left\{(\mathbf{d}_i^{(g)}, y_i^{(g)})\right\}_{g=1}^{G} \sim \pi_{\theta}(\cdot \mid \mathbf{x}_i)
\end{equation}
As shown in Figure~\ref{fig:Overview_RLCS}(b), each output receives a binary reward based on whether its judgment matches the ground truth:
\begin{equation}
    r_i^{(g)} = \mathbb{I}[y_i^{(g)} = y_i^*] \cdot 2 - 1 \in \{-1, +1\}
\end{equation}

Before computing advantages, we apply entropy-based reward shaping (detailed in Section~\ref{sec:Reward_Shaping}) to obtain shaped rewards $\{r_i'^{(g)}\}_{g=1}^{G}$, as depicted in the “Advantage Strategy” component of Figure~\ref{fig:Overview_RLCS}(b). The group-relative advantage is then computed as:
\begin{equation}
    A_i^{(g)} = r_i'^{(g)} - \frac{1}{G}\sum_{g'=1}^{G} r_i'^{(g')}
\end{equation}

The GRPO training objective is:

\begin{align}
    \mathcal{L}_{\text{GRPO}} = -\mathbb{E}_{\mathcal{D}_{\text{RL}}}\Bigg[&\sum_{g=1}^{G} \min\bigg(\rho_i^{(g)} A_i^{(g)}, \notag \\
    &\text{clip}(\rho_i^{(g)}, 1-\epsilon, 1+\epsilon) A_i^{(g)}\bigg)\Bigg] \notag \\
    &+ \beta \cdot \mathbb{E}[\text{KL}(\pi_{\theta} \| \pi_{\text{SFT}})]
\label{eq:grpo}
\end{align}

where the probability ratio $\rho_i^{(g)}$ is defined as:
\begin{equation}
    \rho_i^{(g)} = \frac{\pi_{\theta}(\mathbf{d}_i^{(g)}, y_i^{(g)} \mid \mathbf{x}_i)}{\pi_{\theta_{\text{old}}}(\mathbf{d}_i^{(g)}, y_i^{(g)} \mid \mathbf{x}_i)}.
\end{equation}

The term $\pi_{\theta_{\text{old}}}$ is the policy from the previous iteration, $\pi_{\text{SFT}}$ is the cold-start policy, and $\epsilon, \beta$ control clipping and KL regularization respectively.

Through this two-stage training approach, which combines supervised learning on high-quality demonstrations with reinforcement learning on expanded preference data, we obtain a GenRM capable of generating detailed reasoning that explains story quality across multiple dimensions. This validated GenRM subsequently serves as the reward function for training the story generation policy, as illustrated in Figure~\ref{fig:Overview_RLCS}(c).

\subsection{Entropy-Based Reward Shaping}
\label{sec:Reward_Shaping}
While the GenRM provides rich feedback for response quality, naively applying this reward in reinforcement learning can lead to training instability. The key insight is that not all samples contribute equally to policy improvement. We operationalize a reward shaping strategy that considers both model confidence (measured by output entropy) and response correctness.

For each sample $i$ with generated output group $\{(\mathbf{d}_i^{(g)}, y_i^{(g)})\}_{g=1}^{G}$, we compute the token-level entropy for each output $(g)$ and aggregate it into a trajectory-level entropy $H_i^{(g)}$. Using the batch median $\tau_H = \text{median}(\{H_j^{(g')}\}_{j,g'})$ as an adaptive threshold, we classify samples into four categories and assign different weight multipliers:

\begin{itemize}
    \item \textbf{Low confidence, incorrect} ($H_i^{(g)} > \tau_H$, $r_i^{(g)} = -1$): Standard weight $w_i^{(g)} = 1.0$, as high uncertainty on errors is expected during exploration.
    \item \textbf{High confidence, incorrect} ($H_i^{(g)} \leq \tau_H$, $r_i^{(g)} = -1$): Enhanced weight $w_i^{(g)} = 1.5$, as confident errors expose systematic misconceptions requiring strong correction.
    \item \textbf{Low confidence, correct} ($H_i^{(g)} > \tau_H$, $r_i^{(g)} = +1$): Enhanced weight $w_i^{(g)} = 1.5$, encouraging consolidation of fortuitous correct behaviors.
    \item \textbf{High confidence, correct} ($H_i^{(g)} \leq \tau_H$, $r_i^{(g)} = +1$): Reduced weight $w_i^{(g)} = 0.5$ to prevent overfitting on already-mastered patterns.
\end{itemize}

Formally, the shaped reward is $r_i'^{(g)} = w_i^{(g)} \cdot r_i^{(g)}$, where:
\begin{equation}
w_i^{(g)} = 
\begin{cases}
1.0 & H_i^{(g)} > \tau_H, r_i^{(g)} = -1 \\
1.5 & H_i^{(g)} \leq \tau_H, r_i^{(g)} = -1 \\
1.5 & H_i^{(g)} > \tau_H, r_i^{(g)} = +1 \\
0.5 & H_i^{(g)} \leq \tau_H, r_i^{(g)} = +1
\end{cases}
\end{equation}

This mechanism concentrates learning on confidently incorrect and uncertainly correct samples, while preventing over-optimization on confident correct patterns, leading to more stable and sample-efficient training.

\section{Experiments}
Our experimental evaluation is designed to rigorously validate the RLCS framework through two main parts. First, we evaluate the effectiveness of our GenRM on an expert-annotated test set with pairwise story comparisons, measuring its alignment with human preferences. Second, we demonstrate the effectiveness of the story generation model trained with the validated GenRM as the reward signal. Finally, through a series of ablation studies, we isolate and quantify the contribution of each core component of our framework. Detailed experimental configurations are provided in Appendix~\ref{app:experimental_setup}.

\subsection{Experimental Setup}
\paragraph{Base Models.}
For GenRM training, we employ three base models of dif\-fer\-ent scales: \texttt{Qwen2.5-7B-In\-struct}, \texttt{Qwen2.5-14B-In\-struct}, and \texttt{Qwen2.5-32B-In\-struct}. Each model under\-goes the two-stage training pipeline described in Section~4. For story generation, we ini\-tial\-ize from a \texttt{Qwen-72B} model that has been further pre\-trained on ap\-prox\-i\-mate\-ly 30B tokens of story-related corpora.

\paragraph{Data and Baselines.}
Our evaluation uses an expert-annotated test set of 500 high-quality preference pairs labeled by professional screenwriters. We compare RLCS against: (1) \textbf{SFT-Base}: The Qwen-72B model after initial SFT; (2) \textbf{Standard RL (GRPO w/ Discriminative Reward)}: GRPO with a traditional discriminative Bradley-Terry reward model; (3) \textbf{Gemini-2.5-Pro}: A state-of-the-art commercial model. Additional details on data construction, baselines, and evaluation metrics are in Appendix~\ref{app:experimental_setup}.

\subsection{Validation of the GenRM}
We first evaluate the effectiveness of our two-stage GenRM training pipeline across different model scales. Table~\ref{tab:genrm-comprehensive} presents the accuracy of GenRM models of varying sizes on the expert-annotated evaluation set, comparing both the SFT-only baseline and the full SFT+GRPO pipeline.

Several key observations emerge from these results. First, across all model scales, the GRPO refinement stage provides consistent and substantial improvements over SFT-only models, with gains ranging from $4.9\%$ to $6.3\%$. This validates the effectiveness of our two-stage training approach. Second, larger models achieve higher absolute performance, with the $32B$ model reaching $68.3\%$ accuracy, substantially outperforming the discriminative Bradley-Terry reward model ($54.1\%$). Third, even the smallest $7B$ model after GRPO refinement ($64.5\%$) significantly surpasses the discriminative baseline, demonstrating that our generative approach with explicit reasoning is more effective than discriminative methods regardless of model scale.

Additionally, we compare our best GenRM (32B, SFT+GRPO) against state-of-the-art commercial models acting as judges. Table~\ref{tab:genrm-comprehensive} demonstrates that GenRM achieves 68.3\% consistency agreement with expert annotators, outperforming Gemini-2.5-Pro (60.0\%) and Claude-4-Sonnet (62.0\%). This validates that our specialized reward model, trained specifically for story evaluation through our two-stage pipeline, surpasses even powerful general-purpose models in capturing creative writing preferences.

\begin{table*}[t]
  \centering
  \small
  \caption{Comprehensive GenRM performance comparison across model scales and baselines. All our models benefit substantially from GRPO refinement after SFT, with the best configuration outperforming commercial alternatives.}
  \label{tab:genrm-comprehensive}
  \begin{tabular}{@{}lccccc@{}}
    \toprule
    \textbf{Model} & \textbf{Type} & \textbf{Trained} & \textbf{SFT (\%)} & \textbf{SFT+GRPO (\%)} & \textbf{Gain (\%)} \\
    \midrule
    \multicolumn{6}{l}{\textit{Our GenRM Models}} \\
    \addlinespace[0.5ex]
    Qwen2.5-7B-Instruct  & Generative & \checkmark & 58.2 & 64.5 & +6.3 \\
    Qwen2.5-14B-Instruct & Generative & \checkmark & 61.8 & 66.7 & +4.9 \\
    Qwen2.5-32B-Instruct & Generative & \checkmark & 63.1 & \textbf{68.3} & +5.2 \\
    \midrule
    \multicolumn{6}{l}{\textit{Baseline Models}} \\
    \addlinespace[0.5ex]
    Qwen2.5-7B-Instruct  & Discriminative & \checkmark & -- & 51.8 & -- \\
    Qwen2.5-14B-Instruct & Discriminative & \checkmark & -- & 53.2 & -- \\
    Qwen2.5-32B-Instruct & Discriminative & \checkmark & -- & 54.1 & -- \\
    Gemini-2.5-Pro       & Generative     & --         & -- & 60.0 & -- \\
    Claude-4-Sonnet      & Generative     & --         & -- & 62.0 & -- \\
    \bottomrule
  \end{tabular}
\end{table*}

\subsection{Story Generation Performance with GenRM Guidance}
We now evaluate the effectiveness of using our trained GenRM as the reward signal for story generation. Starting from a Qwen-72B model pretrained on 30B tokens of story-related corpora, we first perform SFT on the story generation dataset (establishing the SFT-Base baseline), then apply GRPO training guided by our GenRM. We conducted a large-scale human evaluation where annotators compared stories generated by RLCS against those from our baselines for 300 unseen prompts.

Table~\ref{tab:generation-results} shows that RLCS is overwhelmingly preferred by human evaluators. It achieves a win rate of $72.4\%$ against the SFT-Base model and $66.8\%$ against the Standard RL baseline trained with a discriminative reward model. Notably, RLCS also outperforms the strong Gemini-2.5-Pro baseline with a win rate of $59.1\%$, demonstrating the effectiveness of our integrated approach combining articulated rewards with entropy-based optimization. 

\begin{table*}[t]
  \centering
  \caption{Head-to-head win rates for story generation. RLCS significantly outperforms all baselines.}
  \label{tab:generation-results}
  \begin{tabular}{lccc}
    \toprule
    \textbf{Comparison} & \textbf{Win (\%)} & \textbf{Tie (\%)} & \textbf{Lose (\%)} \\
    \midrule
    \textbf{RLCS} vs. SFT-Base & \textbf{72.4} & 10.1 & 17.5 \\
    \textbf{RLCS} vs. Standard RL (Discriminative) & \textbf{66.8} & 12.3 & 20.9 \\
    \textbf{RLCS} vs. Gemini-2.5-Pro & \textbf{59.1} & 15.5 & 25.4 \\
    \bottomrule
  \end{tabular}
\end{table*}

To further isolate the contribution of GenRM versus the discriminative reward model, we trained a variant \texttt{RLCS-Discriminative} that uses our stable GRPO optimization and entropy-based reward shaping but is guided by the weaker discriminative reward model instead of GenRM. In a head-to-head comparison, the full \textbf{RLCS model was preferred over \texttt{RLCS-Discriminative} with a win rate of $62.7\%$}. This confirms that the rich, multi-dimensional feedback and explicit reasoning from GenRM are critical for guiding the policy towards high-quality narrative structures, and its benefit extends beyond mere preference prediction accuracy.

\subsection{Ablation Studies}
\begin{figure}[h]
\centering
\includegraphics[width=\columnwidth]{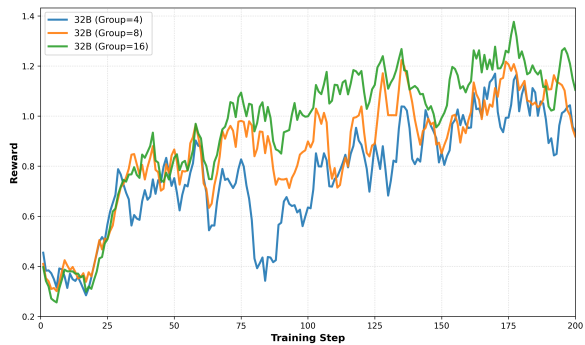}
\caption{Impact of group rollout size on GenRM performance. Performance saturates around $G=8$.}
\label{fig:rollout-ablation}
\end{figure}

\begin{figure*}[!h]
  \centering
  \includegraphics[width=\textwidth]{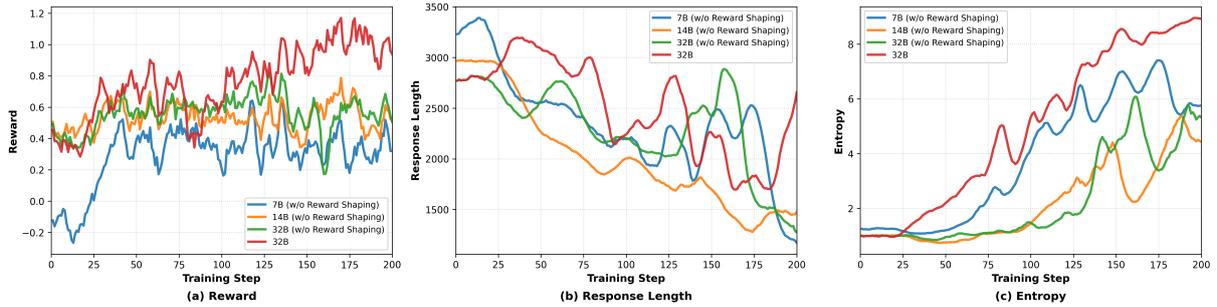}
  \caption{Training dynamics of GenRM across different model scales. The GRPO phase consistently improves upon SFT initialization.}
  \label{fig:training-curve}
\end{figure*}

We conduct ablation studies to isolate the contribution of key components. Figure~\ref{fig:rollout-ablation} shows that moderate rollout diversity ($G=8$) provides optimal performance. Figure~\ref{fig:training-curve} demonstrates the training dynamics across model scales, revealing that larger models achieve higher rewards, learn to generate appropriately-lengthed responses, and maintain exploration through increased entropy. Finally, comparing RLCS with a uniform reward weighting variant (\texttt{RLCS-Uniform}) shows that our entropy-based reward shaping achieves both more stable training and higher final quality (58.9\% win rate), validating its importance for effective policy optimization. Detailed ablation analyses are provided in Appendix~\ref{app:ablation_details}.

\section{Conclusion}
\label{sec:conclusion}

In this paper, we introduced the Reinforcement Learning for Creative Storytelling (RLCS) framework, a comprehensive approach designed to address the critical challenges of reward modeling and training instability in creative story generation. To overcome the limitations of coarse scalar feedback from traditional methods like the BT model, we develop a Generative Reward Model (GenRM) that articulates multi-dimensional analysis of story quality and explicitly reasons about preferences. GenRM is trained through a two-stage pipeline: a supervised fine-tuning cold-start phase ensures task alignment and structured reasoning generation, followed by a RL phase that continuously refines judgment accuracy. To ensure stable and efficient RL training, we introduce an entropy-based reward shaping strategy that dynamically prioritizes learning on samples where the model exhibits confident errors or uncertain correctness, while preventing overfitting on already-mastered patterns.
Our experiments demonstrate the effectiveness of this integrated approach. The trained GenRM achieves 68\% alignment with human creativity judgments, and the complete RLCS framework significantly outperforms strong baselines including Gemini-2.5-Pro in overall story quality. These results validate both the value of articulated rewards for capturing nuanced storytelling preferences and the importance of targeted optimization strategies for stable RL training in creative domains. This work establishes a practical and effective methodology for applying RL to subjective creative domains, demonstrating that careful attention to both reward design and training stability can unlock the potential of LLMs for high-quality creative generation.

\section*{Limitations}
\label{sec:limitations}
While our work demonstrates promising results in RL-based creative story generation, several limitations should be acknowledged. First, the subjective nature of creative storytelling means that our GenRM reflects the preferences of the particular annotators involved in the training process, which may not fully capture all aesthetic perspectives or cultural contexts. Second, our framework has been primarily validated on story generation tasks, and its applicability to other creative domains such as poetry or screenplay writing requires further investigation. Finally, the entropy-based reward shaping strategy, while effective in our experiments, may require task-specific tuning when adapted to different creative generation scenarios.

We acknowledge several risks in deploying creative AI systems. Our GenRM may inherit biases from training data, potentially generating stories that reinforce stereotypes or exclude certain perspectives. The system may also produce factually incorrect, culturally insensitive, or inappropriate content, and could be misused to create deceptive narratives. To mitigate these risks, we strongly recommend mandatory human review before deployment.


\nocite{Ando2005,andrew2007scalable,rasooli-tetrault-2015}

\clearpage 

\bibliography{custom}

\appendix
\section{Training Algorithms}
\label{app:algorithms}
This appendix provides detailed algorithmic descriptions of the two key training procedures in RLCS:
\begin{enumerate}
    \item \textbf{Generative Reward Model (GenRM) Training} (Algorithm~\ref{alg:cot_sft}): Two-stage training that first performs supervised fine-tuning on reasoning chain demonstrations, then applies GRPO with entropy-based reward shaping on preference data to improve reward accuracy
    \item \textbf{Story Generation with GRPO} (Algorithm~\ref{alg:grpo_with_rm}): Policy optimization using the trained GenRM as the reward signal for creative storytelling
\end{enumerate}

\label{app:reward_model}
\begin{algorithm*}[htbp] 
\caption{Generative Reward Model Training}
\label{alg:cot_sft}
\begin{algorithmic}[1]\small
    \State \textbf{Input:} Pretrained LLM $\pi_\theta$; CoT-SFT dataset $D_{SFT} = \{(q_i, c_i, y_i)\}_{i=1}^{N}$; SFT batch size $B$; SFT separator token `[SEP]`; SFT epochs $E_{SFT}$, SFT learning rate $\eta_{SFT}$; GRPO main steps $T_{GRPO}$, GRPO learning rate $\eta_{GRPO}$; on-policy samples number $n$; GRPO update epochs $E_{update}$; rule-based verifier $v$.
    \State \textbf{Output:} Fine-tuned policy $\pi_{\theta}^{*}$
    \Statex \textit{\textcolor{blue} {Phase 1: SFT Regularization}}
    \For{$e=1$ to $E_{SFT}$}
        \State Shuffle $D_{SFT}$
        \For{each batch $\{(q_j, c_j, y_j)\}_{j=1}^{B}$ in $D_{SFT}$}
            \State Initialize batch loss $\mathcal{L}_{batch} \leftarrow 0$
            \For{$j=1$ to $B$}
                \Statex \Comment{Construct the full target sequence by concatenating CoT and the final answer}
                \State Construct target sequence $T_j \leftarrow c_j \oplus \texttt{[SEP]} \oplus y_j$
                \Statex \Comment{Compute standard auto-regressive cross-entropy loss}
                \State Compute loss $\mathcal{L}_j \leftarrow -\sum_{t=1}^{|T_j|} \log \pi_\theta(T_j^{(t)} \mid q_j, T_j^{(<t)})$
                \State $\mathcal{L}_{batch} \leftarrow \mathcal{L}_{batch} + \mathcal{L}_j$
            \EndFor
            \Statex \Comment{Update model parameters based on the average batch loss}
            \State $\theta \leftarrow \theta - \eta_{SFT} \nabla_\theta \left(\frac{1}{B}\mathcal{L}_{batch}\right)$
        \EndFor
    \EndFor
    \State Store the SFT-tuned model: $\pi_{\theta}^{SFT} \leftarrow \pi_{\theta}$.

    \Statex
    \Statex \textit{\textcolor{blue} {Phase 2: GRPO Policy Optimization}}
    \State Initialize policy from SFT-tuned model: $\pi_{\theta} \leftarrow \pi_{\theta_{SFT}}$.
    \For{$t=1$ to $T_{GRPO}$}
        \Statex \Comment{\textit{On-policy Rollout Phase: Collect a batch of experience}}
        \State Initialize rollout buffer $D_{rollout} \leftarrow \emptyset$.
        \State Sample a batch of questions $\{q_j\}_{j=1}^{M}$ and their corresponding supervising trajectories $\{\tau_j^*\}_{j=1}^{M}$ from $D_{SFT}$.
        \For{$j=1$ to $M$}
            \For{$i=1$ to $n$}
                \State Sample trajectory (COT+Answer) $\tau_{j,i} \sim \pi_\theta(\cdot \mid q_j)$ and evaluate its reward $R(\tau_{j,i}) \leftarrow v(\tau_{j,i})$.
                \State Add tuple $(q_j, \tau_j^*, \tau_{j,i}, R(\tau_{j,i}))$ to $D_{rollout}$.
            \EndFor
        \EndFor

        \Statex \Comment{\textit{Multi-epoch Update Phase: Use the collected experience multiple times}}
        \For{$e_{update}=1$ to $E_{update}$} 
            \State Shuffle the collected data $D_{rollout}$.
            \For{each mini-batch from $D_{rollout}$}
                \State Compute on-policy RL loss $\mathcal{L}_{RL}$ on the mini-batch.
                \State Update parameters: $\theta \leftarrow \theta - \eta_{GRPO} \nabla_\theta \mathcal{L}_{RL}$.
            \EndFor
        \EndFor
    \EndFor
    \State \textbf{return} $\pi_{\theta}$ as $\pi_{\theta}^*$.
\end{algorithmic}
\end{algorithm*}

\label{app:grpo}
\begin{algorithm*}[htbp]
\caption{Story Generation with GRPO}
\label{alg:grpo_with_rm}
\begin{algorithmic}[1]\small
    \State \textbf{Input:} SFT-tuned policy model $\pi_{\theta}$; Pretrained generative Reward Model $RM_\phi$; SFT dataset $D_{SFT}$; GRPO main steps $T_{GRPO}$, GRPO learning rate $\eta_{GRPO}$, on-policy samples $n$, GRPO update epochs $E_{update}$.
    \State \textbf{Output:} Final optimized policy $\pi_{\theta}^*$.

    \For{$t=1$ to $T_{GRPO}$}
        \Statex \Comment{\textit{On-policy Rollout Phase}}
        \State Initialize rollout buffer $D_{rollout} \leftarrow \emptyset$.
        \State Sample a batch of questions $\{x_j\}_{j=1}^{M}$ and their corresponding supervising trajectories $\{\tau_j^*\}_{j=1}^{M}$ from $D_{SFT}$.
        \For{$j=1$ to $M$}
            \State \textbf{Policy model generates $n$ trajectories}: $\{\tau_{j,1}, \dots, \tau_{j,n}\} \sim \pi_\theta(\cdot \mid x_j)$.
            \Statex \Comment{\textit{Pairwise Reward Calculation using a Random Pivot}}
            \State Randomly select a pivot index $p \in \{1, \dots, n\}$. Let $\tau_{pivot} \leftarrow \tau_{j,p}$.
            \For{$i=1$ to $n$}
                \If{$i = p$}
                    \State $R(\tau_{j,i}) \leftarrow 0$. 
                \Else
                    \State $R(\tau_{j,i}) \leftarrow RM_\phi(\tau_{j,i}, \tau_{pivot})$. 
                \EndIf
                \State Add tuple $(x_j, \tau_j^*, \tau_{j,i}, R(\tau_{j,i}))$ to $D_{rollout}$.
            \EndFor
        \EndFor

        \Statex \Comment{\textit{Multi-epoch Update Phase}}
        \For{$e_{update}=1$ to $E_{update}$}
            \State Shuffle $D_{rollout}$ and iterate through its mini-batches.
            \For{each mini-batch from $D_{rollout}$}
                \State Compute adaptive parameters $\alpha, \beta$ based on the rewards in the mini-batch.
                \State Compute on-policy RL loss $\mathcal{L}_{RL}$ and SFT loss $\mathcal{L}_{SFT}$.
                \State Combine losses: $\mathcal{L} \leftarrow \alpha \mathcal{L}_{RL} + \beta \mathcal{L}_{SFT}$.
                \State Update policy parameters: $\theta \leftarrow \theta - \eta_{GRPO} \nabla_\theta \mathcal{L}$. \Comment{RM parameters $\phi$ are frozen}
            \EndFor
        \EndFor
    \EndFor
    \State \textbf{return} $\pi_{\theta}$ as $\pi_{\theta}^*$.
\end{algorithmic}
\end{algorithm*}

\section{Detailed Experimental Setup}
\label{app:experimental_setup}

\subsection{Preference Data for GenRM Training}
The preference data for training both our GenRM and baseline discriminative reward models are sourced from machine annotations generated by large language models such as Gemini-2.5-Pro and Claude-4-Sonnet. Following best practices in preference learning, we employ multi-model consensus to ensure annotation quality, where multiple strong models vote on story preferences and only high-agreement samples are retained for training. This process yields high-quality preference pairs that combine human judgments (used after filtering for SFT) and synthetic consensus data (used for GRPO refinement).

\subsection{Story Generation Training Data}
For the story generation task, each training instance consists of three input components and one target output: (1) \textbf{Story Background}: contextual setting and premise; (2) \textbf{Story Outline}: high-level plot structure; (3) \textbf{Character Profiles}: descriptions of key characters and their traits; and (4) \textbf{Target Story}: a complete story excerpt (up to 2,000 characters) extracted from real published novels. To construct this dataset, we employ a reverse-engineering approach using Gemini-2.5-Pro and other strong models to distill the background, outline, and character profiles from authentic literary works through carefully designed prompts. This process yields approximately \textbf{5,000 high-quality training instances} covering diverse genres and narrative styles.

\subsection{Pairwise to Pointwise Reward Conversion}
Since GenRM is trained for pairwise comparisons, we adopt a reference-based strategy to convert them into pointwise rewards following \citep{jia2025writingzerobridgegapnonverifiable}. For each group rollout, we randomly sample one response as the reference and construct pairs with all other responses in the group. GenRM evaluates each pair and assigns a pointwise reward of $+1$ to responses preferred over the reference and $-1$ to inferior ones, providing the reward signal required for policy optimization.

\subsection{Expert-Annotated Evaluation Set}
Recognizing that standard automatic metrics fail to capture the nuanced quality of creative writing, we construct a high-quality, expert-annotated evaluation dataset designed specifically to assess preference alignment in story generation. Each instance is a tuple $(C, S_A, S_B)$, comprising a rich \textbf{Narrative Context ($C$)} and a pair of \textbf{Candidate Stories ($S_A, S_B$)}. These story pairs are deliberately designed to differ in key creative dimensions such as dialogue quality, narrative pacing, and plot coherence. To ensure professional standards, we employ expert screenwriters for annotation. After reviewing the context, annotators provide a binary preference for each story pair along with detailed qualitative rationales justifying their choices based on criteria such as coherence, character consistency, and dramatic impact. This meticulous process results in a final evaluation set of \textbf{500 high-quality, expert-annotated preference pairs}.

\subsection{Evaluation Metrics}
Our evaluation consists of two parts. For \textit{reward model evaluation}, we measure \textbf{Accuracy (\%)} on the expert-annotated evaluation set, defined as the percentage of times a model's preference matches expert judgments. For \textit{story generation evaluation}, we conduct human assessments measuring head-to-head \textbf{Win Rate (\%)} comparing RLCS against baselines, and collect absolute ratings on a $1\sim12$ scale for \textbf{Coherence}, \textbf{Creativity}, and \textbf{Engagement}.

\section{Detailed Ablation Studies}
\label{app:ablation_details}

\subsection{Impact of Group Rollout Size}
We investigate how the number of rollout samples $G$ during GRPO training affects GenRM's final performance. Figure~\ref{fig:rollout-ablation} in the main text shows the validation accuracy across different rollout sizes. We observe that performance improves as $G$ increases from $4$ to $16$, with diminishing returns beyond $G=8$. This suggests that moderate diversity in rollout samples ($8\sim16$) provides sufficient signal for advantage estimation without excessive computational cost. We use $G=8$ as the default setting in our main experiments.

\subsection{Training Dynamics of GenRM}
To understand the learning behavior of our two-stage training pipeline, we monitor key metrics during the GRPO phase across different model scales (Figure~\ref{fig:training-curve} in the main text). All models start from the same SFT checkpoint.

Figure~\ref{fig:training-curve}(a) shows that larger models achieve higher and more stable rewards, with the 32B model (red) reaching approximately $1.0$ while smaller models (7B/14B, blue/orange) plateau at approximately $0.3\sim0.5$ with greater oscillation. Figure~\ref{fig:training-curve}(b) reveals an interesting emergent behavior: models learn to reduce response length from initial $3000+$ tokens to $1500\sim2500$ tokens, discovering that quality storytelling does not require excessive verbosity. Figure~\ref{fig:training-curve}(c) demonstrates that reward shaping helps maintain exploration, with entropy steadily increasing to $8\sim9$ for the 32B model, preventing premature convergence.

These dynamics validate our design: SFT provides stable initialization, while GRPO enables continued improvement through reward-guided exploration. The 32B model's superior performance across all metrics confirms its selection as our final GenRM.

\subsection{Impact of Entropy-Based Reward Shaping}
To isolate the benefit of our entropy-based reward shaping strategy, we trained a variant named \texttt{RLCS-Uniform}. This model uses the same GenRM reward signal and GRPO algorithm as our full framework but applies uniform reward weighting to all training samples, rather than dynamically prioritizing confident errors and uncertain correct predictions. We performed two analyses. First, during training, the \texttt{RLCS-Uniform} model exhibited significantly higher reward variance and occasional instabilities, whereas our full \texttt{RLCS} model showed a smooth and monotonically increasing reward curve with lower variance. Second, in a head-to-head human evaluation, the full \textbf{RLCS model was preferred over \texttt{RLCS-Uniform} with a win rate of 58.9\%}. These results collectively demonstrate that our entropy-based reward shaping strategy is crucial not only for making the training process more stable and efficient but also for achieving higher final story quality by focusing learning on the most informative samples.

\section{Prompts for Data Synthesis}
\label{sec:appendix_2}
\begin{figure*}[!htbp]
    \centering
    \includegraphics[width=\textwidth]{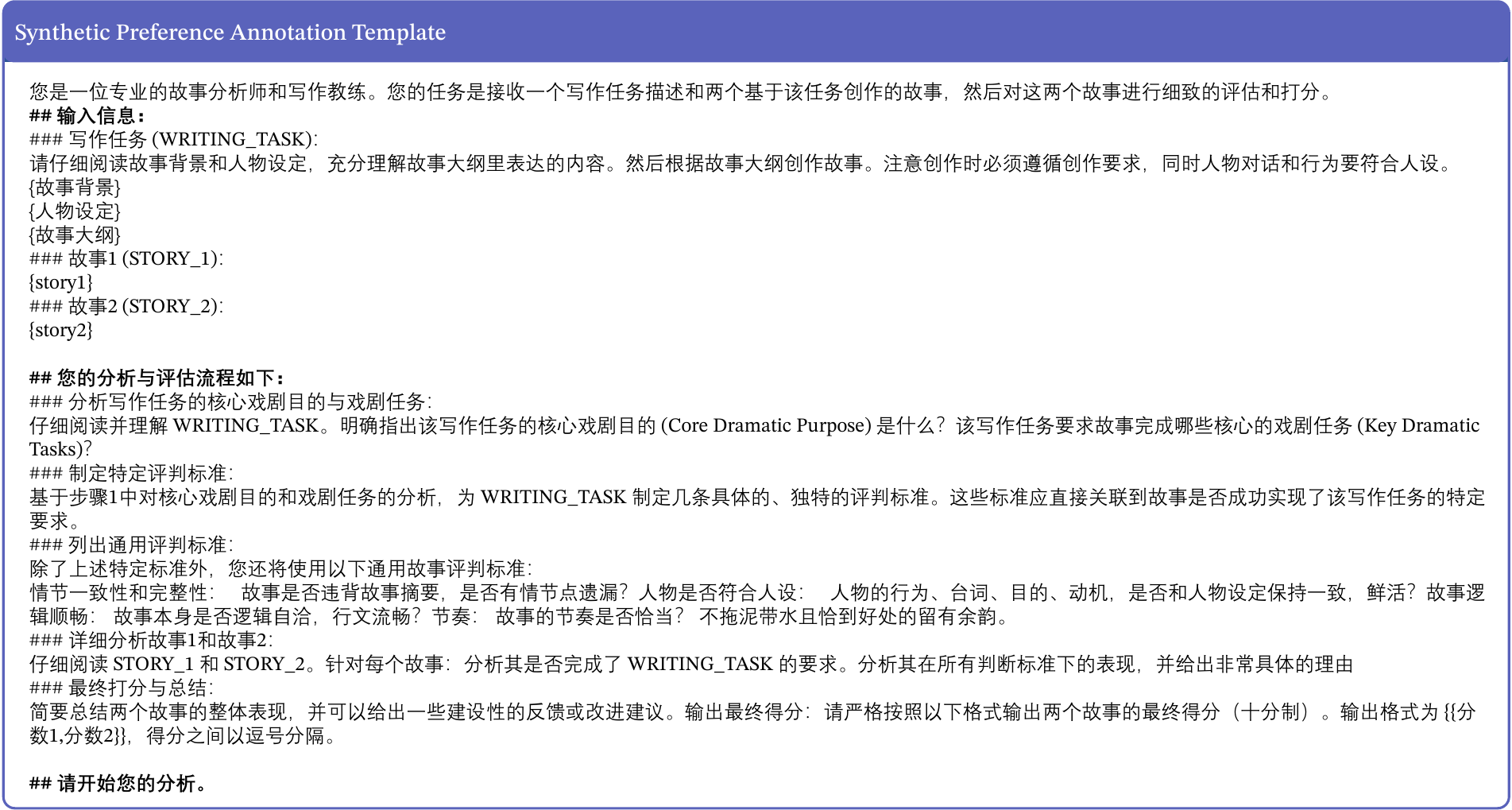} 
    \caption{Prompt template for automated story preference labeling.}
    \label{fig:preference labeling}
\end{figure*}

\begin{figure*}[!htbp]
    \centering
    \includegraphics[width=\textwidth]{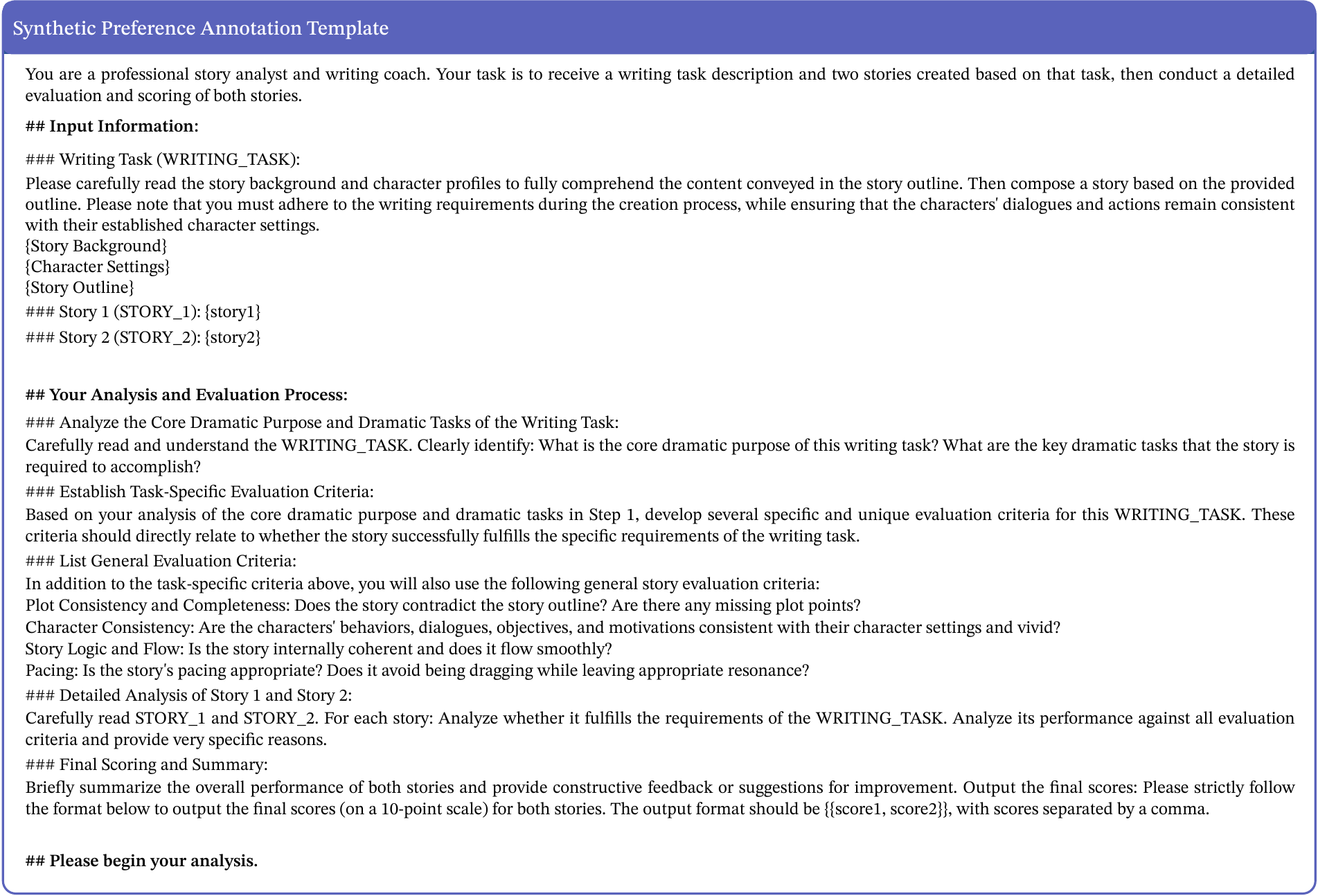} 
    \caption{Prompt template for automated story preference labeling translated into English.}
    \label{fig:preference labeling}
\end{figure*}

\end{document}